\documentclass[runningheads]{llncs}
\usepackage[T1]{fontenc}
\usepackage{graphicx,verbatim}
\usepackage{amssymb, amsmath}
\usepackage[capitalise]{cleveref}
\usepackage{subcaption}
\usepackage{multicol,multirow}
\usepackage{threeparttable}
\begin{document}
\title{Spatial-Aware Self-Supervision for Medical 3D Imaging with Multi-Granularity Observable Tasks}

\begin{comment}  %% Removed for anonymized MICCAI 2025 submission
    \author{First Author\inst{1}\orcidID{0000-1111-2222-3333} \and
    Second Author\inst{2,3}\orcidID{1111-2222-3333-4444} \and
    Third Author\inst{3}\orcidID{2222--3333-4444-5555}}
    %
    \authorrunning{F. Author et al.}
    \institute{Princeton University, Princeton NJ 08544, USA \and
    Springer Heidelberg, Tiergartenstr. 17, 69121 Heidelberg, Germany
    \email{lncs@springer.com}\\
    \url{http://www.springer.com/gp/computer-science/lncs} \and
    ABC Institute, Rupert-Karls-University Heidelberg, Heidelberg, Germany\\
    \email{\{abc,lncs\}@uni-heidelberg.de}}
    
\end{comment}

\author{Yiqin Zhang, Meiling Chen, Zhengjie Zhang}  %% Added for anonymized MICCAI 2025 submission
\authorrunning{Zhang, et al.}
\institute{University of Shanghai for Science and Technology \\
    \email{zyqmgam@163.com}}

\titlerunning{Spatial-Aware Self-Sup}
\maketitle
\begin{abstract}
    The application of self-supervised techniques has become increasingly prevalent within medical visualization tasks, primarily due to its capacity to mitigate the data scarcity prevalent in the healthcare sector. The majority of current works are influenced by designs originating in the generic 2D visual domain, which lack the intuitive demonstration of the model’s learning process regarding 3D spatial knowledge. Consequently, these methods often fall short in terms of medical interpretability. We propose a method consisting of three sub-tasks to capture the spatially relevant semantics in medical 3D imaging. Their design adheres to observable principles to ensure interpretability, and minimize the performance loss caused thereby as much as possible. By leveraging the enhanced semantic depth offered by the extra dimension in 3D imaging, this approach incorporates multi-granularity spatial relationship modeling to maintain training stability. Experimental findings suggest that our approach is capable of delivering performance that is on par with current methodologies, while facilitating an intuitive understanding of the self-supervised learning process.

\end{abstract}

\section{Introduction}

Self-supervised learning has long been a favorite among researchers in medical visual domains, as it aids in overcoming the high cost of medical annotations, thereby enabling contemporary neural networks to achieve superior performance on downstream tasks~\cite{VanBerlo2024}. Typically characterized by CT and MRI, 3D imaging and analysis in the medical field are hindered by barriers such as high information volume, complex semantics, and non-mainstream modalities~\cite{Upadhyay2024}.

Both CT and MRI are methods of visualizing the internal tissues and organs of the human body. While the specifics of each imaging sample can vary, they all have overarching semantic structures in common, like the heart, stomach, and kidneys. Therefore, it is feasible to train a neural network to recognize the 3D distribution of internal human organs by analyzing a large number of samples~\cite{VanBerlo2024}. Contrastive learning has become a very popular self-supervised learning method in recent years. VoCo~\cite{Wu_2024_CVPR}, CVRL~\cite{You10.1007} are aimed at migrating contrastive learning techniques from the broad visual domain to the specific realm of 3D medical imaging, yielding promising outcomes. Another classic technique is Masked Image Modeling (MIM), multiple researches~\cite{Chen_2023_WACV,Tian10.1007} have applied it to medical imaging domains.

Humans naturally understand 3D space easily, so visualizing the model’s learning process of 3D spatial knowledge would help promote the development of trustworthy medical AI~\cite{NAZIR2023106668}. However, in the current research on medical self-supervision, there is still a lack of relevant studies~\cite{Zeng2024BioMedical}, with most research tending to transfer the 2D development paradigm to 3D and relatively few considerations on the additional dimension.

In this study, we involves extracting multiple sub-regions from a volume and employing three self-supervised tasks to model the spatial position semantics within both the physical coordinate system and the latent space: \textbf{1)} As a coarse-grained method, Coupled Relative Similarity Classification ensures basic proximity judgment; \textbf{2)} As a medium granularity method, Gap Matrix Prediction roughly obtains similarity semantics; And \textbf{3)} as a fine-grained approach, Route-Based Connectivity Supervision to learn detailed spatial body structure knowledge. The method enables us to acquire comprehensive semantic knowledge regarding the entire volume. The computational complexity of our method is independent of the volume size. It operates by sampling a predetermined number of sub-regions within the volume, thereby incurring low peak memory demands. 

Compared to other methods, the approach we present incorporates similarity measures to compel the model to learn the spatial arrangement of human tissues. The supervision derived from graph routes provides a straightforward means of ascertaining the model’s capability to accurately discern the appropriate positions of various tissues within the human body. This is valuable for augmenting the interpretability of self-supervised techniques in the medical field.

It must be acknowledged that although our design exhibits certain parallels with GMIM \cite{GMIM2024108547}, there are notable distinctions. Their methodology necessitates a limited degree of annotation, incorporates more overt learning and focus objectives, and lacks the capability to obtain supervisory information at varying levels of granularity. Our study diverges from theirs in these regards.

\section{Method}

\subsection{Overview of Spatial-Aware Design}

\begin{figure}[tb]
    \centering
    \includegraphics[width=\textwidth]{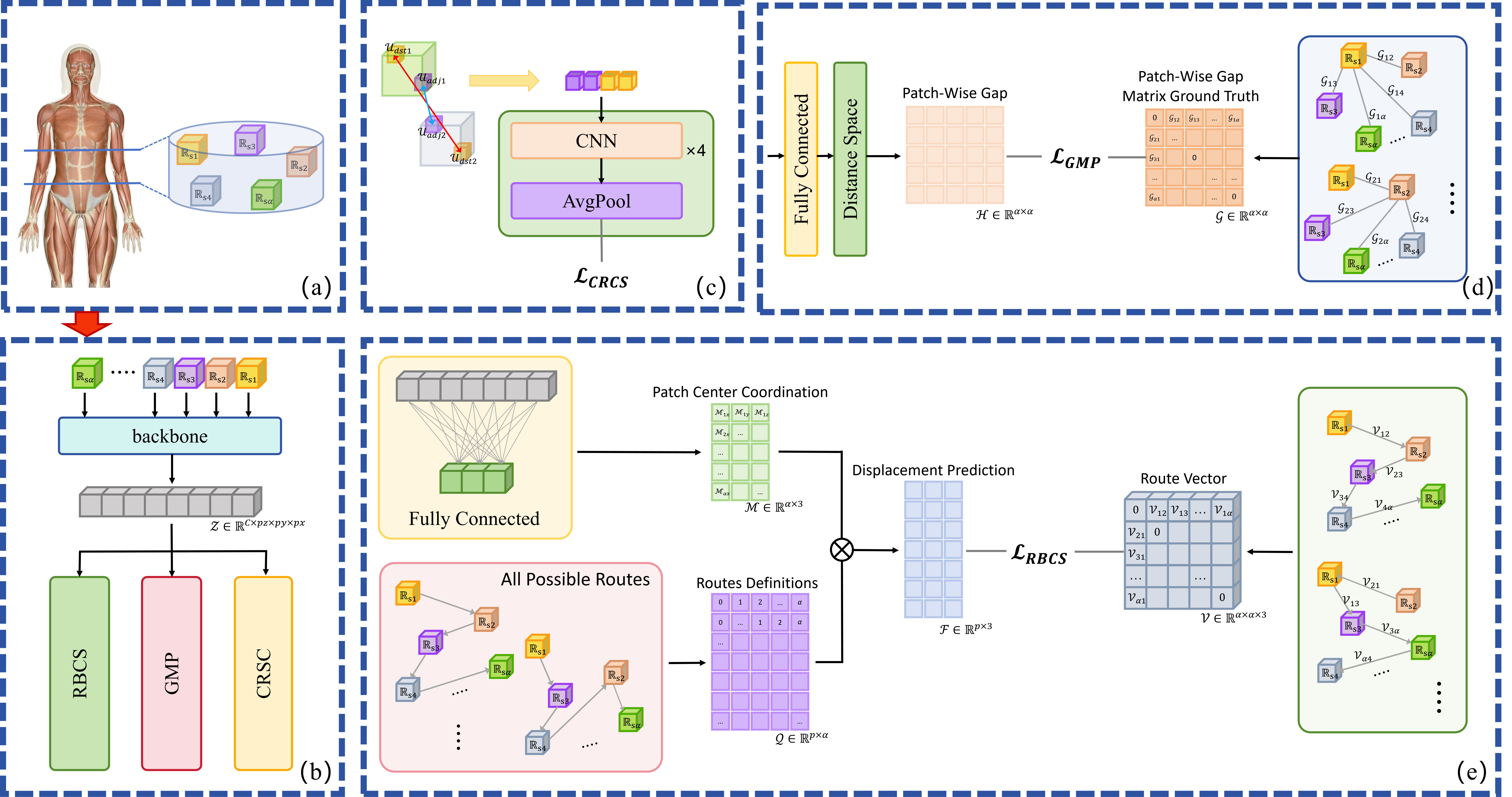}
    \caption{Overview of the proposed spatial-aware self-supervised learning. (c): Coupled Relative Similarity Classification, (d): Gap Matrix Prediction, (e): Route-Based Connectivity Supervision.}
    \label{fig:overview}
\end{figure}

Our method includes a shared backbone network and three distinct self supervised task: 1) CRSC - \textbf{C}oupled \textbf{R}elative \textbf{S}imilarity \textbf{C}lassification, 2) GMP - \textbf{G}ap \textbf{M}atrix \textbf{P}rediction, and 3) RBCS - \textbf{R}oute-\textbf{B}ased \textbf{C}onnectivity \textbf{S}uperviser. Collectively, these components derive spatial similarity semantics $\mathcal{S}$ from an extensive array of unlabelled medical 3D volumes $\mathbb{R}^{z \times y \times x}$, utilizing the acquired insights to inform subsequent tasks that are sensitive to spatial dynamics. In \cref{fig:overview}, we provide a general overview of how they learn from one another and the elements that constitute the ultimate self-supervised goal.

In general, we partition a Volume into $\alpha$ patches. These patches are independently processed by the backbone for feature extraction. To ensure the optimization stability of the backbone, only the forward pass of a randomly selected patch is involved in gradient computation, while the remaining patches undergo gradient-free inference using a momentum-averaged model, thereby obtaining the neural latent space representations of the patches. The three self-supervised subtasks are then trained based on these representations.

\subsection{Coupled Relative Similarity Classification}
\label{sec:relsim}

Our design incorporates multi-granularity self-supervision, necessitating a stable and non-varying supervisory goal. Predicting the absolute similarity within different areas $\mathbb{R}_{s1}, \mathbb{R}_{s2} \in \mathbb{R}^{pz \times py \times px}$ of a 3D volume $\mathbb{R}$ is tough, given that the target range spans a wide continuum from complete irrelevance (-1) to full correlation (1). Consequently, we opt to resample the closest and furthest (antipodal) volume pair within $\mathbb{R}_{s1}, \mathbb{R}_{s2}$. We use $\mathcal{U} = \{\mathbb{R}_{adj1}, \mathbb{R}_{adj2}, \mathbb{R}_{dst1}, \mathbb{R}_{dst2}~\vert~\mathbb{R}^{vz \times vy \times vx}\}$ to mark the closest and furthest pairs, and it is the basic sample unit for training in this section.

A simple cosine similarity neural calculation module is deployed to calculate the similarity matrix $\mathbb{J} \in \mathbb{R}^{4 \times 4}$ between all possible sub-regions combinition among $\mathcal{U}$. Noted that the volume data requires pre-extraction using the shared backbone, so calculation of $\mathbb{J}$ is performed in the latent space. The loss criterion is designed to maximize similarity of $\mathcal{U}_{adj}$ and minimize between $\mathcal{U}_{dst}$, as is described as $\mathcal{L}_{\text{CRSC}} = \cos(\mathcal{U}_{dst}) - \cos(\mathcal{U}_{adj})$. During inferencing, the group of subdomains with smaller cosine similarity is considered as $\mathcal{U}_{dst}$.

With more than two $\mathbb{R}_s$ present, a pair of them can always produce one $\mathcal{U}$ that can be learned from. Assuming the number of sub-volumes of $\mathbb{R}_s$ is $\alpha$, the shape of the similarity matrix will be $\mathbb{J} \in \mathbb{R}^{\mathbf{C}^2_\alpha  \times 4 \times 4}$. It is apparent that the self-supervision semantics equals to $\mathcal{O}(\alpha^2)$ , whereas the computational demand is $\mathcal{O}(\alpha)$. Consequently, we are empowered to judiciously select larger $\alpha$ to intensify self-supervision without significant apprehension regarding computational constraints.

\subsection{Gap Matrix Prediction}
\label{sec:gap}

Next, we use the second prediction branch to produce the latent space coordinates for each Rs. These coordinates are then mapped back to the world coordinate system, giving us the model’s predicted positions for each $\mathbb{R}_s$. A major hurdle here is the near-impossibility of creating a consistent coordinate system across various scan outcomes. This is because when imaging each subject, factors such as the starting and ending points of imaging, body posture, and physique are determined by specific clinical diagnostic needs and equipment conditions. There is significant variation between samples, making it impossible to determine the origin of the world coordinate system that is applicable for all volumes.

Consequently, we have adopted a strategy of predicting absolute spatial position gap. Data adhering to the DICOM standard is reconstructed incorporating a spacing parameter, which accurately reflects the dimensional information of each voxel in the physical realm. This enables us to ascertain the physical gaps between each volume, thereby utilizing these intervals as the self-supervised objective. The positions $\mathcal{P}_s \in \mathbb{R}^{3}$ corresponding to each $\mathbb{R}_s$ are determined via a direct output mapping from latent space. With $\alpha$ sub-volumes, we can calculate the gap matrix $\mathcal{G} \in \mathbb{R}^{\alpha \times \alpha}$, where $\mathcal{G}_{ij} = {\Vert \mathcal{P}_i - \mathcal{P}_j \Vert}_2$. Therefore, the loss is calculated as $\mathcal{L}_{GMP}=\frac{1}{\alpha^2}\sum_{i,j=1}^{\alpha}(\frac{\vert \widehat{\mathcal{G}}_{ij} - \mathcal{G}_{ij} \vert}{\mathcal{G}_{ij} + \epsilon})^2$

Upon extensive training, the neural network’s output converges to a biased world coordinate system, as the bias in the coordinate system does not impact the computation of the gap loss and thus remains uncorrected by the supervisory mechanism. However, the variability in anatomical dimensions among subjects and the imprecise alignment of organ positions across different individuals mean that corresponded sub-regions from two people do not possess a consistent gap. This reduces the upper limit of the model's fitting ability in this part. In view of this, we reduced the granularity of supervision, allowing the network to focus only on the L2 distance between $\mathbb{R}_s$, without calculating the loss for the distance difference in each dimension separately.

\subsection{Route-Based Connectivity Supervision}
\label{sec:route}

In \cref{sec:gap}, we implemented absolute distance supervision between $\mathbb{R}_s$ and decreased the level of detail to guarantee the consistency of the fitting process. In this section, we introduce an additional module designed to mitigate the semantic loss incurred by the reduction in detail.

Similarly, we first use a branch head to map the latent space to the world coordinate system. The central coordinates of all $\mathbb{R}_s$ are represented by the set $\mathcal{M} \in \mathbb{R}^{\alpha \times 3}$. Then, a depth-first backtracking algorithm is deployed to search all possible traversals through each element in $\mathcal{M}$, and these traversals are recorded as $\mathcal{Q} \in \mathbb{R}^{p \times \alpha}$, where $p$ denotes the number of traversals and the order is represented by the index of $\mathbb{R}_s$. For path $\mathcal{Q}_k=(s_1 \rightarrow s_2 \rightarrow \cdots \rightarrow s_\alpha)$ generated via depth-first search, compute cumulative displacement error $\triangle_k = {\parallel \sum_{t=1}^{\alpha-1} [ (\widehat{\mathcal{M}}_{s_{t+1}} - \widehat{\mathcal{M}}_{s_{t}}) - (\mathcal{M}_{s_\alpha} - \mathcal{M}_{s_1}) ] \parallel}_2 $, then the loss is calculated as $\mathcal{L}_{RBCS} = \frac{1}{K} \sum_{k=1}^K \triangle_k^2$.

The design employs a 3D path accumulation technique to forecast the distance between sub-regions, as opposed to deriving the similarity measure directly from the neural implicit space. This method can be perceived as integrating spatial location priors, thereby facilitating the model’s awareness of the relational positioning of all sub-regions. Throughout the training phase, the visualization of path predictions provides a tangible indicator of the neural network’s acquisition of effective spatial location knowledge.

\section{Experiments}

\subsection{Datasets and Baselines}

In light of their representativeness and practicality, we have elected to train the MedNeXt~\cite{Roy2023MedNeXt} and SegFormer3D~\cite{PereraSegFormer3D} networks, which are extensively recognized within the academic and clinical communities. Then, MoCoV3~\cite{Chen_2021_ICCV} and SimMIM~\cite{Xie9880205} are deployed as contrastive and reconstruction baselines. Our training process exclusively employs publicly accessible datasets, as detailed in \cref{tab:datasets}. All volumes are resampled and patched to $96 \times 96 \times 96mm$ for any further processing, and the window level is $200HU$, window width is $800HU$ for CT modality. We utilize a foreground filtering strategy to decrease the sampling rate of empty areas in imaging, which boosts learning efficiency. As a result, the number of Instances and Patches doesn't have a fixed correlation across different datasets.

\begin{table}[tb]
    \centering
    \caption{Datasets used in the experiments.}
    \label{tab:datasets}
    \begin{tabular}{lcc}
        \hline
        Dataset & Instances & \shortstack{Pretrained\\Patches} \\
        \hline
        TotalSeg.~\cite{dantonoli2024totalsegmentator} & 1228 & 275780 \\
        FLARE2023~\cite{FLARE2023} & 4000 & 1462948 \\
        KiTS23~\cite{heller2023kits21} & 489 & 143300 \\
        AbdomenCT1K~\cite{MaAbdomenCT1K} & 1062 & 264852 \\
        LUNA16~\cite{SETIO20171} & 888 & 150456 \\
        \hline
    \end{tabular}
\end{table}

\subsection{Pretraining}

\begin{figure}[tb]
    \centering
    \begin{subfigure}{0.32\linewidth}
        \includegraphics[width=\linewidth]{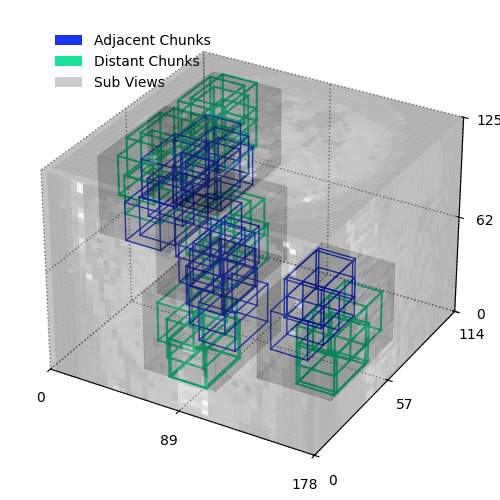}
        \caption{Rel. Sim. Cls.}
        \label{fig:exp_relsim}
    \end{subfigure}
    \begin{subfigure}{0.32\linewidth}
        \includegraphics[width=\linewidth]{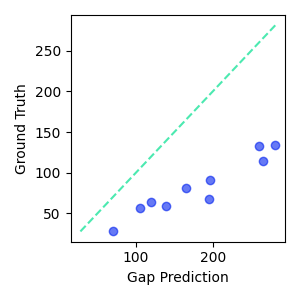}
        \caption{Gap Matrix Pred.}
        \label{fig:exp_gap}
    \end{subfigure}
    \begin{subfigure}{0.32\linewidth}
        \includegraphics[width=\linewidth]{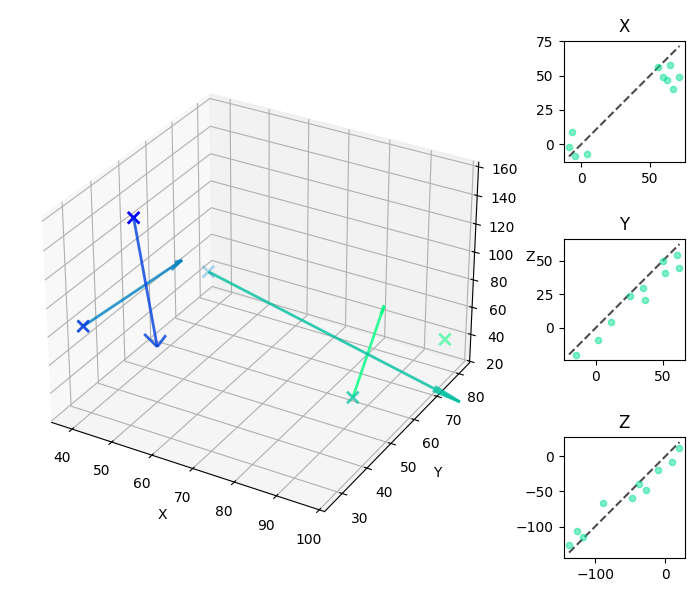}
        \caption{Route-Based C. S.}
        \label{fig:exp_route}
    \end{subfigure}
    \caption{Visualization of the performance on self-supervision tasks.}
    \label{fig:exp_selfsup}
\end{figure}

CRSC (\cref{sec:relsim}) shows excellent accuracy in \cref{fig:exp_relsim}. It can serve as a good supervisory target, guiding the neural network to distinguish the spatial proximity relationships of different human body regions, and thereby learn relevant semantic knowledge. When $\alpha$ is significant, it becomes challenging to visually classify similarities because of the overwhelming number of pairs. However, a straightforward and intuitive assessment is to observe whether adjacent pairs are more concentrated towards the central area of the volume. The rationale behind this is that the calculated distance between the central point coordinates of adjacent pairs and the volume’s central point follows an expectation of $\{(s - p + v)/2~\vert~s \in \{z,y,x\}, p \in \{pz, py, px\}, v \in \{vz, vy, vx\}\}$, in contrast to $(s + p - v)/2$ for distant pairs. Given $p > v$, adjacent pairs are inherently more proximate to the volume’s center. 

In analyzing the GMP (\cref{sec:gap}), a scatter plot (\cref{fig:exp_gap}) is employed to visualize the discrepancy between predicted and actual outcomes. It reveals that while the model retains some error in specific predictions, an overall trend of convergence is apparent.

The RBCS (\cref{sec:route}) shows the similar trend in \cref{fig:exp_route}. It is important to highlight that, during training, we aggregate all vectors along a specific path before applying supervision. The depiction provided here, where each vector originates from its intended starting point, offers a more accurate representation of the error contributions from individual segments.

\begin{figure}[tb]
    \centering
    \begin{subfigure}{0.32\linewidth}
        \includegraphics[width=\linewidth]{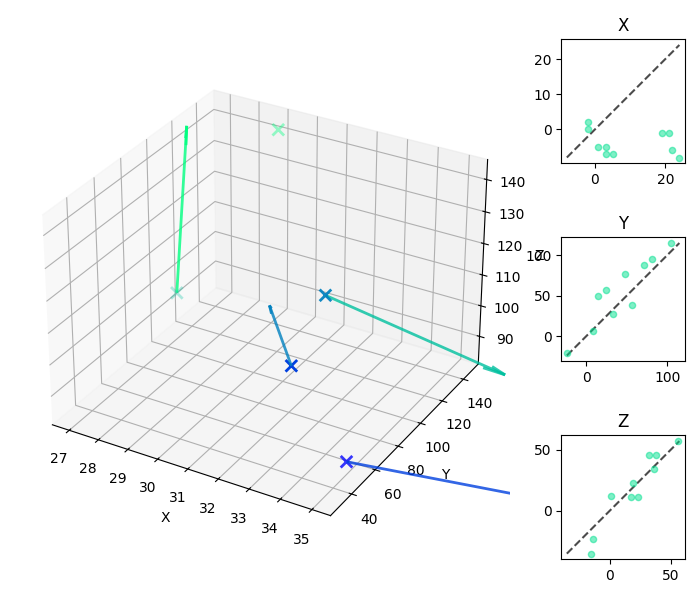}
        \caption{Iter. 5000}
        \label{fig:exp_RouteVis1}
    \end{subfigure}
    \begin{subfigure}{0.32\linewidth}
        \includegraphics[width=\linewidth]{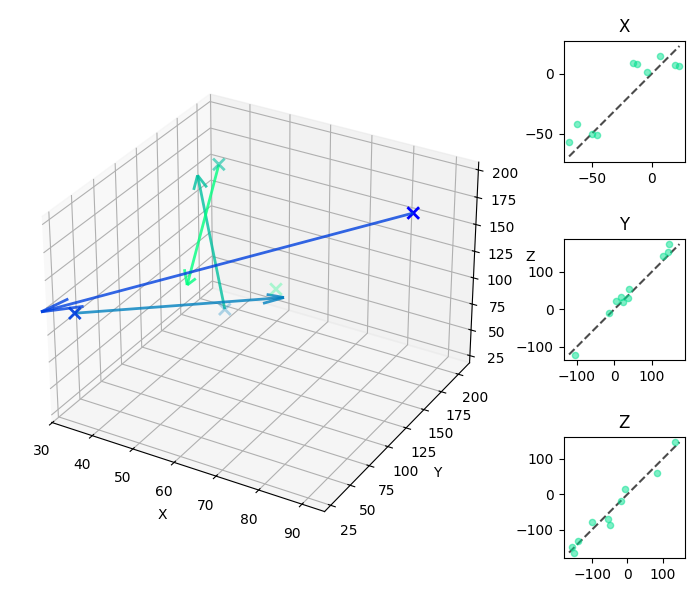}
        \caption{Iter. 25000}
        \label{fig:exp_RouteVis2}
    \end{subfigure}
    \begin{subfigure}{0.32\linewidth}
        \includegraphics[width=\linewidth]{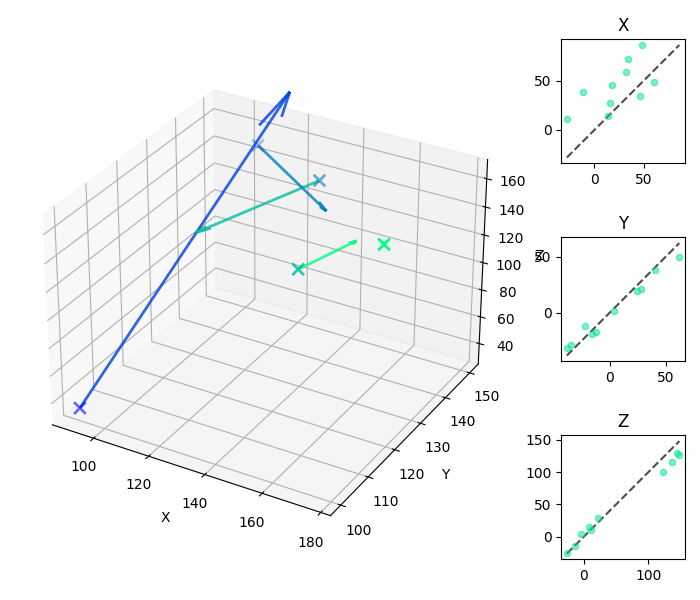}
        \caption{Iter. 70000}
        \label{fig:exp_RouteVis3}
    \end{subfigure}
    \caption{Route Connectivity Observation during training.}
    \label{fig:exp_RouteVis}
\end{figure}

As is shown in \cref{fig:exp_RouteVis}, through the examination of visualized results across diverse training stages, we are able to intuitively ascertain the model’s comprehension of the spatial positioning of various tissues within the human body, thereby ensuring the reliability of its predictions. In the absence of comprehensive self-learning, the model faces challenges in establishing the positional correlations between sub-regions. This is evident when the arrows fail to align with the 'X' symbols that correspond to their respective colors.

\subsection{Segmentation Accuray}

The process of fine-tuning a pre-trained model for specific downstream tasks serves as a dependable methodology for the quantitative evaluation of the extent to which the proposed approach successfully extracts semantic information from extensive data. The pre-training phase involved 100K updates to the parameters, followed by an additional 50K updates during the fine-tuning stage. The fine-tuning costs about 35.5 RTX 4090 GPU hours for MedNeXt and 5.6 hours for SegFormer3D.

\cref{tab:segmentation} shows the quantitative evaluation of the segmentation tasks using different training strategies. The prevalent self-supervised methodologies demonstrate sufficient accuracy in the AbdomenCT1K segmentation task. Furthermore, our proposed method also exhibits comparable efficacy when the supervision strategy is distinctly and visually articulated. However, it is observed that all designs incorporating self-supervision with fine-tuning still lag behind those employing full supervision. With the backbone frozen, the compact SegFormer3D model finds it challenging to perform well on downstream tasks, and fails to segment pancreas which is the hardest target of AbdomenCT1K dataset. Considering the SegFormer3D originally has limited learnable parameters, the freeze operation may further limit its ability to segment difficult organs. In contrast, the more substantial MedNeXt network, wich much more parameters, manages to enhance its learning efficiency under similar conditions.
\begin{table}
    \centering
    \begin{threeparttable}
        \caption{Segmentation accuracy with different strategies.}
        \label{tab:segmentation}
        \begin{tabular}{ll|ccccc}
            \hline
            Model & \shortstack{Pretrain\\Method} & Dice & IoU & Recall & Prec. & \shortstack{Pretrain\\Time}\tnote{$\ddagger$} \\
            \hline
            \multirow{5}{*}{MedNeXt~\cite{Roy2023MedNeXt}} & w/o & 94.20 & 89.61 & 93.18 & 95.30 & N/A \\
            & MoCoV3 & 91.25 & 85.23 & 87.49 & 96.46 & 62.3 \\
            & SimMIM & 93.55 & 88.58 & 91.88 & 95.60 & 16.4 \\
            & Ours & 91.39 & 85.57 & 88.76 & 94.88 & \multirow{2}{*}{75.1} \\
            & Ours\tnote{$\dagger$} & 92.23 & 86.75 & 89.74 & 95.37 & \\
            \hline
            \multirow{5}{*}{SegFormer3D~\cite{PereraSegFormer3D}} & w/o & 91.74 & 86.05 & 89.87 & 94.14 & N/A \\
            & MoCoV3 & 91.04 & 85.03 & 89.05 & 93.71 & 9.0 \\
            & SimMIM & 89.30 & 82.52 & 85.91 & 93.93 & 4.1 \\
            & Ours & 90.84 & 84.95 & 88.61 & 94.03 & \multirow{2}{*}{74.7} \\
            & Ours\tnote{$\dagger$} & 63.80 & 55.00 & 58.05 & 92.37 & \\
            \hline
        \end{tabular}
        \begin{tablenotes}
            \item[$\dagger$] Freeze Backbone.
            \item[$\ddagger$] RTX 4090 GPU Hours.
        \end{tablenotes}
    \end{threeparttable}
\end{table}

\section{Discussion and Conclusions}

The explicit modeling of spatial positional relationships facilitates an efficient observation of the neural network’s process in capturing spatial positions, thereby confirming its capability to discern various anatomical regions of the human body during training. Our research proves the feasibility of developing explicit and visualization-centric self-supervised designs within the healthcare domain, without detriment to performance. Further research is essential to meticulously analyze the manner in which neural networks acquire knowledge about human tissue distribution, thereby striking a balance between interpretability and learning efficiency.

While the majority of existing methodologies do not explicitly incorporate spatial correlations in their modeling, the inability to delineate the precise mechanisms through which neural networks acquire knowledge precludes the confirmation of whether these methods inherently attend to spatial correlation during self-supervised learning processes. For instance, methodologies based on MIM~\cite{Xing10416665} are probable to internalize a diversity of specific local voxel paradigms. When inferencing, these methods utilize a limited subset of voxel characteristics to approximate spatial positioning, and subsequently, they tend to yield outputs that represent the anatomical configuration of organs at the inferred locations within the human body~\cite{ijcai2023p762}.

\begin{comment}  %% removed for anonymized MICCAI 2025 submission.
    
    % The following acknowledgement and disclaimer sections should be removed for the double-blind review process.  
    % If and when your paper is accepted, reinsert the acknowledgement and the disclaimer clause in your final camera-ready version.

\begin{credits}
\subsubsection{\ackname} A bold run-in heading in small font size at the end of the paper is
used for general acknowledgments, for example: This study was funded
by X (grant number Y).

\subsubsection{\discintname}
It is now necessary to declare any competing interests or to specifically
state that the authors have no competing interests. Please place the
statement with a bold run-in heading in small font size beneath the
(optional) acknowledgments\footnote{If EquinOCS, our proceedings submission
system, is used, then the disclaimer can be provided directly in the system.},
for example: The authors have no competing interests to declare that are
relevant to the content of this article. Or: Author A has received research
grants from Company W. Author B has received a speaker honorarium from
Company X and owns stock in Company Y. Author C is a member of committee Z.
\end{credits}

\end{comment}

\bibliographystyle{splncs04}
\bibliography{SpatialAwareSelfSup.bib}

\begin{thebibliography}{10}
\providecommand{\url}[1]{\texttt{#1}}
\providecommand{\urlprefix}{URL }
\providecommand{\doi}[1]{https://doi.org/#1}

\bibitem{Chen_2021_ICCV}
Chen, X., Xie, S., He, K.: An empirical study of training self-supervised vision transformers. In: Proceedings of the IEEE/CVF International Conference on Computer Vision (ICCV). pp. 9640--9649 (October 2021)

\bibitem{Chen_2023_WACV}
Chen, Z., Agarwal, D., Aggarwal, K., Safta, W., Balan, M.M., Brown, K.: Masked image modeling advances 3d medical image analysis. In: Proceedings of the IEEE/CVF Winter Conference on Applications of Computer Vision (WACV). pp. 1970--1980 (January 2023)

\bibitem{dantonoli2024totalsegmentator}
D'Antonoli, T.A., Berger, L.K., Indrakanti, A.K., Vishwanathan, N., Weiß, J., Jung, M., Berkarda, Z., Rau, A., Reisert, M., Küstner, T., Walter, A., Merkle, E.M., Segeroth, M., Cyriac, J., Yang, S., Wasserthal, J.: Totalsegmentator mri: Sequence-independent segmentation of 59 anatomical structures in mr images (2024), \url{https://arxiv.org/abs/2405.19492}

\bibitem{heller2023kits21}
Heller, N., Isensee, F., Trofimova, D., Tejpaul, R., Zhao, Z., Chen, H., Wang, L., Golts, A., Khapun, D., Shats, D., Shoshan, Y., Gilboa-Solomon, F., George, Y., Yang, X., Zhang, J., Zhang, J., Xia, Y., Wu, M., Liu, Z., Walczak, E., McSweeney, S., Vasdev, R., Hornung, C., Solaiman, R., Schoephoerster, J., Abernathy, B., Wu, D., Abdulkadir, S., Byun, B., Spriggs, J., Struyk, G., Austin, A., Simpson, B., Hagstrom, M., Virnig, S., French, J., Venkatesh, N., Chan, S., Moore, K., Jacobsen, A., Austin, S., Austin, M., Regmi, S., Papanikolopoulos, N., Weight, C.: The kits21 challenge: Automatic segmentation of kidneys, renal tumors, and renal cysts in corticomedullary-phase ct (2023)

\bibitem{FLARE2023}
Ma, J., Wang, B. (eds.): Fast, Low-resource, and Accurate Organ and Pan-cancer Segmentation in Abdomen CT. Lecture Notes in Computer Science, Springer Cham, 1 edn. (2024). \doi{10.1007/978-3-031-58776-4}

\bibitem{MaAbdomenCT1K}
Ma, J., Zhang, Y., Gu, S., Zhu, C., Ge, C., Zhang, Y., An, X., Wang, C., Wang, Q., Liu, X., Cao, S., Zhang, Q., Liu, S., Wang, Y., Li, Y., He, J., Yang, X.: Abdomenct-1k: Is abdominal organ segmentation a solved problem? IEEE Transactions on Pattern Analysis and Machine Intelligence  \textbf{44}(10),  6695--6714 (Oct 2022). \doi{10.1109/TPAMI.2021.3100536}

\bibitem{NAZIR2023106668}
Nazir, S., Dickson, D.M., Akram, M.U.: Survey of explainable artificial intelligence techniques for biomedical imaging with deep neural networks. Computers in Biology and Medicine  \textbf{156},  106668 (2023). \doi{https://doi.org/10.1016/j.compbiomed.2023.106668}, \url{https://www.sciencedirect.com/science/article/pii/S0010482523001336}

\bibitem{PereraSegFormer3D}
Perera, S., Navard, P., Yilmaz, A.: Segformer3d: an efficient transformer for 3d medical image segmentation. In: 2024 IEEE/CVF Conference on Computer Vision and Pattern Recognition Workshops (CVPRW). pp. 4981--4988 (June 2024). \doi{10.1109/CVPRW63382.2024.00503}

\bibitem{GMIM2024108547}
Qi, L., Jiang, Z., Shi, W., Qu, F., Feng, G.: Gmim: Self-supervised pre-training for 3d medical image segmentation with adaptive and hierarchical masked image modeling. Computers in Biology and Medicine  \textbf{176},  108547 (2024). \doi{https://doi.org/10.1016/j.compbiomed.2024.108547}

\bibitem{Roy2023MedNeXt}
Roy, S., Koehler, G., Ulrich, C., Baumgartner, M., Petersen, J., Isensee, F., J{\"a}ger, P.F., Maier-Hein, K.H.: Mednext: Transformer-driven scaling of convnets for medical image segmentation. In: Greenspan, H., Madabhushi, A., Mousavi, P., Salcudean, S., Duncan, J., Syeda-Mahmood, T., Taylor, R. (eds.) Medical Image Computing and Computer Assisted Intervention -- MICCAI 2023. pp. 405--415. Springer Nature Switzerland, Cham (2023)

\bibitem{SETIO20171}
Setio, A.A.A., Traverso, A., {de Bel}, T., Berens, M.S., van~den Bogaard, C., Cerello, P., Chen, H., Dou, Q., Fantacci, M.E., Geurts, B., van~der Gugten, R., Heng, P.A., Jansen, B., {de Kaste}, M.M., Kotov, V., Lin, J.Y.H., Manders, J.T., Sóñora-Mengana, A., García-Naranjo, J.C., Papavasileiou, E., Prokop, M., Saletta, M., Schaefer-Prokop, C.M., Scholten, E.T., Scholten, L., Snoeren, M.M., Torres, E.L., Vandemeulebroucke, J., Walasek, N., Zuidhof, G.C., van Ginneken, B., Jacobs, C.: Validation, comparison, and combination of algorithms for automatic detection of pulmonary nodules in computed tomography images: The luna16 challenge. Medical Image Analysis  \textbf{42},  1--13 (2017), \url{https://www.sciencedirect.com/science/article/pii/S1361841517301020}

\bibitem{Tian10.1007}
Tian, Y., Pang, G., Liu, Y., Wang, C., Chen, Y., Liu, F., Singh, R., Verjans, J.W., Wang, M., Carneiro, G.: Unsupervised anomaly detection in medical images with a memory-augmented multi-level cross-attentional masked autoencoder. In: Cao, X., Xu, X., Rekik, I., Cui, Z., Ouyang, X. (eds.) Machine Learning in Medical Imaging. pp. 11--21. Springer Nature Switzerland, Cham (2024)

\bibitem{Upadhyay2024}
Upadhyay, A.K., Bhandari, A.K.: Advances in deep learning models for resolving medical image segmentation data scarcity problem: A topical review. Archives of Computational Methods in Engineering  \textbf{31}(3),  1701--1719 (04 2024). \doi{10.1007/s11831-023-10028-9}

\bibitem{VanBerlo2024}
VanBerlo, B., Hoey, J., Wong, A.: A survey of the impact of self-supervised pretraining for diagnostic tasks in medical x-ray, ct, mri, and ultrasound. BMC Medical Imaging  \textbf{24}(1), ~79 (04 2024). \doi{10.1186/s12880-024-01253-0}

\bibitem{Wu_2024_CVPR}
Wu, L., Zhuang, J., Chen, H.: Voco: A simple-yet-effective volume contrastive learning framework for 3d medical image analysis. In: Proceedings of the IEEE/CVF Conference on Computer Vision and Pattern Recognition (CVPR). pp. 22873--22882 (June 2024)

\bibitem{Xie9880205}
Xie, Z., Zhang, Z., Cao, Y., Lin, Y., Bao, J., Yao, Z., Dai, Q., Hu, H.: Simmim: a simple framework for masked image modeling. In: 2022 IEEE/CVF Conference on Computer Vision and Pattern Recognition (CVPR). pp. 9643--9653 (June 2022). \doi{10.1109/CVPR52688.2022.00943}

\bibitem{Xing10416665}
Xing, Z., Zhu, L., Yu, L., Xing, Z., Wan, L.: Hybrid masked image modeling for 3d medical image segmentation. IEEE Journal of Biomedical and Health Informatics  \textbf{28}(4),  2115--2125 (April 2024). \doi{10.1109/JBHI.2024.3360239}

\bibitem{You10.1007}
You, C., Zhao, R., Staib, L.H., Duncan, J.S.: Momentum contrastive voxel-wise representation learning for semi-supervised volumetric medical image segmentation. In: Wang, L., Dou, Q., Fletcher, P.T., Speidel, S., Li, S. (eds.) Medical Image Computing and Computer Assisted Intervention -- MICCAI 2022. pp. 639--652. Springer Nature Switzerland, Cham (2022)

\bibitem{Zeng2024BioMedical}
Zeng, X., Abdullah, N., Sumari, P.: Self-supervised learning framework application for medical image analysis: a review and summary. BioMedical Engineering OnLine  \textbf{23}(1), ~107 (2024). \doi{10.1186/s12938-024-01299-9}, \url{https://doi.org/10.1186/s12938-024-01299-9}

\bibitem{ijcai2023p762}
Zhang, C., Zhang, C., Song, J., Yi, J.S.K., Kweon, I.S.: A survey on masked autoencoder for visual self-supervised learning. In: Elkind, E. (ed.) Proceedings of the Thirty-Second International Joint Conference on Artificial Intelligence, {IJCAI-23}. pp. 6805--6813. International Joint Conferences on Artificial Intelligence Organization (8 2023). \doi{10.24963/ijcai.2023/762}, survey Track

\end{thebibliography}

\end{document}